\documentclass[runningheads]{llncs}
\usepackage[T1]{fontenc}
\usepackage{graphicx}
\usepackage{color}
\usepackage{subcaption}
\usepackage[english]{babel}
\usepackage[utf8]{inputenc}
\usepackage{algorithm}
\usepackage{arevmath}     
\usepackage[noend]{algpseudocode}
\usepackage{amsmath}
\usepackage{booktabs,dcolumn}
\usepackage{hyperref}
\begin{document}
\title{Election of Collaborators via Reinforcement Learning for Federated Brain Tumor Segmentation}
\titlerunning{Reinforcement Learning for Federated Brain Tumor Segmentation }
%
\author{Muhammad Irfan Khan\inst{1} \and
Elina Kontio\inst{1} \and
Suleiman A. Khan\inst{1} \and
Mojtaba Jafaritadi\inst{1}}
\authorrunning{Khan et al., 2024}
%
\institute{Turku University of Applied Sciences, Turku 20520, Finland 
\\
\email{irfan.khan, elina.kontio, suleiman.alikhan, mojtaba.jafaritadi@turkuamk.fi}
} 
\maketitle              
\begin{abstract}
Federated learning (FL) enables collaborative model training across decentralized datasets while preserving data privacy. However, optimally selecting participating collaborators in dynamic FL environments remains challenging. We present RL-HSimAgg, a novel reinforcement learning (RL) and similarity-weighted aggregation (simAgg) algorithm using harmonic mean to manage outlier data points. This paper proposes applying multi-armed bandit algorithms to improve collaborator selection and model generalization. By balancing exploration-exploitation trade-offs, these RL methods can promote resource-efficient training with diverse datasets. We demonstrate the effectiveness of Epsilon-greedy (EG) and upper confidence bound (UCB) algorithms for federated brain lesion segmentation. In simulation experiments on internal and external validation sets, RL-HSimAgg with UCB collaborator outperformed the EG method across all metrics, achieving higher Dice scores for Enhancing Tumor (0.7334 vs 0.6797), Tumor Core (0.7432 vs 0.6821), and Whole Tumor (0.8252 vs 0.7931) segmentation. Therefore, for the Federated Tumor Segmentation Challenge (FeTS 2024), we consider UCB as our primary client selection approach in federated Glioblastoma lesion segmentation of multi-modal MRIs. In conclusion, our research demonstrates that RL-based collaborator management, e.g. using UCB, can potentially improve model robustness and flexibility in distributed learning environments, particularly in domains like brain tumor segmentation.  

\end{abstract}
%
%
Federated learning (FL) represents a decentralized and democratic framework for collaborative machine learning (ML). {Under this framework, various independent entities with unique datasets and compute resources collaborate without sharing data. These entities, referred to as \textit{collaborators}, function as autonomous units, safeguarding the integrity of their data while working towards a common objective }\cite{mcmahan2017communication,li2020federated}. Just as groups come together for collective decision-making, collaborators in FL engage in a cooperative effort to train a shared ML model. However, unlike centralized ML frameworks, FL is characterized by a distributed nature, akin to a network of independent entities. Collaborators maintain local control over their private or sensitive data and training processes, reflecting their independence within the federated framework. An FL system encourages collaboration and representation, similar to systems that prioritize participation. {Collaborators are chosen for training rounds based on their past contributions to model accuracy, efficiency, and overall improvement. This feedback can influence their selection for future rounds, ensuring that the most effective collaborators are prioritized.} Furthermore, the aggregation of model updates resembles a collective consensus-building process, as collaborators contribute their insights to shape the collective model. This flexible approach accommodates differences in data, similar to variations in preferences across different entities. A cornerstone of FL's systems is balancing the needs of each collaborator with the objective of improving the model as a whole. FL facilitates collaboration and equitable participation, encourages the exchange of knowledge, and advances ML while protecting the privacy of data \cite{Kairouz2019,li2023revisiting,li2020federated}.

Efficiently selecting collaborators for federation rounds presents a complex challenge \cite{shi2021federated,ribero2020communication,cho2020client}. Traditionally, a batch-wise approach has been the standard protocol, where available collaborators are preselected for each round of FL \cite{chen2019communication}. However, this approach often encounters limitations in adapting to dynamic scenarios where new collaborators may join with evolving data distributions and hardware configurations \cite{ali2022federated}. This limitation has motivated the exploration of alternative methods, such as multi-armed bandit algorithms, to enhance the collaborator selection process \cite{kim2020accurate,ali2024communication}. Optimal model training and improved generalization on unseen data can be achieved through performance-based selection of the best possible collaborators. To realize this objective, candidate profiles of collaborators are maintained in each FL round, and collaborators are selected using reinforcement learning (RL), employing a Markov decision process \cite{wu2023fedab}. This introduces healthy competition among collaborators, ensuring effective collaboration. In healthcare and other dynamic fields, the federated learning landscape is constantly evolving. New potential collaborators, each with unique data distributions and hardware setups, frequently join the network. This diversity is often driven by the influx of new patient data at the start of federation rounds, making adaptability a key feature of the system \cite{xu2021federated}. In this context, the application of multi-armed bandit algorithms appears well-suited and adaptable to address this challenge. Here, each collaborator can be likened to a "bandit," and the process of selecting the most promising collaborator can be framed as an optimization problem. The criteria for making this selection can be fine-tuned by striking a balance between exploration and exploitation, effectively managed through a hyperparameter. This hyperparameter can guide the training of the master model at the aggregator, offering the flexibility needed to optimize the selection process. One notable advantage of this approach is its ability to mitigate or eliminate system-level bias in collaborator selection during federation rounds, which is often associated with the automatic grouping of collaborators in batch-wise techniques. 

This work highlights the application of RL for collaborator selection in federated brain tumor segmentation. We leverage Epsilon-greedy (EG) and Upper Confidence Bound (UCB) algorithms to optimize collaborator selection, aiming to improve the efficiency and effectiveness of the FL process. This approach represents a novel application of RL in the domain of brain lesion segmentation, potentially leading to significant advancements in this crucial clinical imaging field.

\section{Methods}
\label{sec:method}

\subsection{Data}
The current study utilized real-world multi-parametric magnetic resonance imaging (mpMRI) data from glioblastoma (GBM) cases made publicly available through the Federated Tumor Segmentation (FeTS) 2022 challenge. GBM is a clinically aggressive brain malignancy classified as a grade IV astrocytoma \cite{hanif2017glioblastoma}. The dataset encompassed 1251 mpMRI scans from confirmed GBM patients, including native T1-weighted images, post-gadolinium contrast enhanced T1-weighted images, T2-weighted images, and T2 Fluid Attenuated Inversion Recovery (FLAIR) images. These cases were derived from the Brain Tumor Segmentation (BraTS) Continuous Challenge dataset, which aims to promote the development and benchmarking of state-of-the-art algorithms for consistent brain tumor delineation across institutions. Specifically, the GBM cohort examined in this study represented a subset of diffuse glioma cases from BraTS containing ground truth segmentations of tumor sub-regions (edema, enhancing core, and necrotic core). By utilizing real-world training data from FeTS and BraTS, we were able to evaluate federated learning for robust GBM segmentation under controlled conditions with known ground truth labels~\cite{baid2021rsna,bakas2017segmentationGBM,bakas2017segmentationLGG,bakas2017advancing,menze2014multimodal}.

The mpMRI data utilized in this study was sourced from multiple academic institutions and preprocessed according to standard protocols including rigid registration, brain extraction, alignment, resolution resampling to uniform voxel sizes, and skull stripping for isolation of intracranial content. To develop and evaluate a federated learning approach for collaborative training of a segmentation model without sharing patient data, we implemented Intel's Federated Learning OpenFL framework along with a U-Net convolutional neural network (CNN) architecture provided through the FeTS 2022 challenge. The federated learning methodology employed a randomly selected subset of participating sites to train local models and update the global model in each round - approximately 20\% of total sites actively contributed to the global model training in a given round. This ensured that no one site had perpetual control over model updates, promoting decentralized collaboration.

\subsection{Multi-Armed Bandit Problem}
In a stochastic multi-armed bandit problem, an agent must choose between $K$ actions ("arms") in order to maximize expected cumulative rewards over time. The reward distributions corresponding to each arm are initially unknown. Multi-armed bandits balance exploring uncertain options and exploiting the currently best option.

\subsubsection{Epsilon-Greedy Strategy}
In the epsilon-greedy strategy, the agent explores arms randomly with a probability of $\varepsilon$, and exploits the best-known arm with a probability of $1 - \varepsilon$. This is done to balance between exploration (finding better arms) and exploitation (gaining rewards from known good arms).

Let $Q(a)$ be the estimated value of arm $a$, and $N(a)$ be the number of times arm $a$ has been pulled. The algorithm can be described as follows:

\begin{enumerate}
    \item Initialize $Q(a)$ and $N(a)$ for all arms.
    \item Repeat for each trial:
    \begin{itemize}
        \item With probability $\varepsilon$, select a random arm ($a = \text{rand\_int } (1, K)$, where $K$ is the number of arms).
        \item Otherwise, select the arm with the highest estimated value: $a = \arg\max_a Q(a)$.
        \item Pull arm $a$, receive reward $R$, and update estimates:
        \begin{align*}
            N(a) &\leftarrow N(a) + 1 \\
            Q(a) &\leftarrow Q(a) + \frac{1}{N(a)} \cdot (R - Q(a))
        \end{align*}
    \end{itemize}
\end{enumerate}

\subsubsection{Upper Confidence Bounds (UCB) Strategy}

The UCB strategy balances exploration and exploitation by choosing arms with the potential to have high rewards based on their upper confidence bounds. This ensures that arms that have been pulled less are given a chance to be explored more.

Let $t$ be the current trial, and $C$ be a parameter that controls the exploration-exploitation trade-off. The UCB algorithm can be outlined as follows:

\begin{enumerate}
    \item Initialize $Q(a)$ and $N(a)$ for all arms.
    \item Repeat for each trial:
    \begin{itemize}
        \item Select the arm that maximizes the upper confidence bound:
        \[
            a = \arg\max_a \left( Q(a) + C \cdot \sqrt{\frac{\ln(t)}{N(a)}} \right)
        \]
        \item Pull arm $a$, receive reward $R$, and update estimates ($N(a)$ and $Q(a)$ as in epsilon-greedy).
    \end{itemize}
\end{enumerate}

In the UCB equation, the term $\sqrt{\frac{\ln(t)}{N(a)}}$ represents the exploration bonus, which decreases with the number of times arm $a$ has been pulled ($N(a)$) and increases with the number of trials ($t$).

\subsection{Election of Collaborators}
\subsubsection{Epsilon-Greedy Approach}

In our federated lesion segmentation framework, the log maintains the validation performance history of all collaborators across rounds. The Epsilon-greedy algorithm is applied for collaborator selection, with a fixed $\varepsilon$ = 0.8 balancing exploration and exploitation. Specifically, a random number is generated each round - if below the exploitation rate of 0.2, the top 20\% highest-performing collaborators from the logs are selected to optimize model performance. But if above 0.2, the bottom 20\% lowest-performing collaborators are chosen to encourage exploration. This alternating selection approach allows both leveraging the strengths of high-performing collaborators and discovering the benefits of underutilized collaborators over the federated learning lifecycle. Like epsilon-greedy bandit algorithms, the federated framework exploits known rewards while still exploring uncertain options, harmonizing the collective learning process.
\begin{algorithm}
\caption{Collaborator Selection using the EG Approach}
\label{alg:collaborator_selection_greedy_epsilon}

\textbf{Input:}
\begin{itemize}
    \item \textit{log}: List of collaborators' validation performance.
    \item \textit{exploitation\_rate}: Proportion of collaborators to exploit (e.g., 0.2 for 20\%).
\end{itemize}

\textbf{Procedure:}
\begin{algorithmic}[1]
    \State Set \textit{num\_to\_select} = \textit{length(log)} $\times$ \textit{exploitation\_rate}.
    \State Generate random number \textit{random\_num} between 0 and 1.
    
    \If{\textit{random\_num} $<$ \textit{exploitation\_rate}}
        \State Sort collaborators by performance in decreasing order.
    \Else
        \State Sort collaborators by performance in increasing order.
    \EndIf
    
    \State Select the top \textit{num\_to\_select} collaborators from the sorted list.
\end{algorithmic}

\textbf{Output:}
\begin{algorithmic}[1]
    \State \Return \textit{selected\_collaborators}.
\end{algorithmic}
\end{algorithm}






This collaborator election approach is well-suited for real-world scenarios as it accommodates scenarios where new collaborators join the federation, and the data distribution undergoes constant changes due to the inclusion of new patient data within a collaborator's domain. The Epsilon-Greedy approach not only ensures equitable participation of all collaborators but also contributes to the optimization of the model's generalization capabilities.

\subsubsection{Upper Confidence Bounds Approach}
In the UCB approach for collaborator selection, a log stores the validation performance history of all collaborators across federation rounds, including the average validation score.

The UCB score for each collaborator is calculated as 
the observed mean reward (validation score) of all available colabs and an exploration bonus proportional to the standard deviation of their observed rewards. This exploration bonus encapsulates uncertainty about collaborators. 

Distinct selection strategies are employed in alternating rounds. In even rounds, the top 20\% of collaborators by highest UCB score are chosen to maximize expected performance. In odd rounds, the bottom 20\% are selected to explore uncertain collaborators.

The selected collaborators then participate in training the global model during the subsequent round. This alternating approach balances leveraging highly rewarding collaborators with exploring uncertain options, providing a principled way to optimize the collective learning process. Over time as uncertainty about collaborator distributions decreases, larger UCB exploration bonuses help identify new high-performing candidates as the federated system evolves.
\begin{algorithm}
\caption{Collaborator Selection using the UCB Approach}
\label{alg:collaborator_selection_ucb}

\textbf{Input:}
\begin{itemize}
    \item \textit{log}: A list of collaborators' validation performance from previous rounds.
\end{itemize}

\textbf{Procedure:}
\begin{algorithmic}[1]
    \State Calculate the average validation score $\text{avg\_score}$.
    
    \For{each collaborator $i$}
        \State Compute the absolute distance from the average score: 
        \State \quad $d_i = |s_i - \text{avg\_score}|$
    \EndFor
    
    \State Sort collaborators based on $d_i$:
    \If{\textit{round\_number} is even}
        \State Sort in increasing order.
    \Else
        \State Sort in decreasing order.
    \EndIf
    
    \State Select the top 20\% of collaborators.
\end{algorithmic}

\textbf{Output:}
\begin{algorithmic}[1]
    \State \Return Selected collaborators for the next round.
\end{algorithmic}
\end{algorithm}



    
    
    
    


\subsection{Similarity Weighted Aggregation}
\label{m1}
In an earlier publication, we introduced a unique aggregation technique known as Similarity Weighted Aggregation (SimAgg). SimAgg efficiently combines model parameters at the server, catering to both Independent Identically Distributed (IID) and non-IID data scenarios, as discussed in \cite{KhanSimAgg}. In this current study, we introduce the RL-HSimAgg algorithm. RL-HSimAgg processes 118 tensors with name containing weight/bias out of 318 total tensors within each federation round, the remaining are processed by FedAvg. It employs RL to determine collaborator selections and utilizes the Harmonic mean for the final aggregation of model parameters. This approach enables effective handling of outliers or instances of extreme values within the dataset.

 \subsubsection{Weight Aggregation Policy}
This section revisits the concept of SimAgg. The core issue when dealing with non-IID data is the possible divergence of model parameters contributed by collaborators. To tackle this challenge, we utilize a weighted aggregation method at the server. Collaborators are assigned weights based on their similarity to the unweighted average. This simple yet effective mechanism enables the creation of a master model that captures the characteristics of the majority of collaborators in each iteration.
The RL-HSimAgg aggregation policy is explained in detail in Algorithm~\ref{simagg_algo}.

During round $r$, the server receives the parameters $p_{C^r}$ contributed by the collaborating entities $C^r$. Subsequently, the server computes the average of these parameters as follows

\begin{equation}
\hat{p} = \frac{1}{\lvert C^r \lvert}\Sigma_{i\in C^r}{p_i}.
\label{eqn_pavg}
\end{equation}

Next, we proceed to determine the inverse distance (similarity) of each collaborator $c$ within $C^r$ from the calculated average

\begin{equation}
sim_c = \frac{\Sigma_{i \in C^r}{\lvert p_i - \hat{p} \rvert}}{\lvert p_c - \hat{p} \rvert + \epsilon},
\label{eqn_sim}
\end{equation}
where $\epsilon = 1e-5$ (small positive constant). 
We standardize the distances to derive the "similarity weights" in the subsequent manner

\begin{equation}
u_c = \frac{sim_c}{\Sigma_{i\in C^r}{sim_i}}.
\label{eqn_sim_weights}
\end{equation}

Collaborators whose parameters closely align with the average are assigned greater similarity weights, whereas those with more significant deviations receive comparatively lower weights. This methodology can effectively mitigate the influence of outliers or instances of substantial divergence, reducing their impact on the aggregation process.

To accommodate the varying influence of distinct sample sizes across each collaborator $c$  within $C^r$, we employ "sample size weights" that prioritize collaborators with a greater number of samples.

\begin{equation}
v_c = \frac{N_c}{\Sigma_{i\in C^r}{N_i}},
\label{eqn_sample_weights}
\end{equation}
where $N_c$ is the number of examples at collaborator $c$. 

Using the weights obtained using Eqs.~\ref{eqn_sim_weights} and~\ref{eqn_sample_weights}, the {\em aggregation weights} are computed as:

\begin{equation}
w_c = \frac{u_c+v_c}{\Sigma_{i\in C^r}{(u_i+v_i)}},
\label{eqn_fsim_weights}
\end{equation}

Ultimately, the aggregation of parameters is done through the harmonic mean of the aggregation weights. The harmonic mean, a statistical average that accommodates situations with reciprocal relationships, is harnessed to consolidate the parameters effectively. This choice of aggregation method enables the model to consider the contributions of collaborators more holistically, especially in scenarios where divergent or extreme values might be present in the dataset. The harmonic mean operates by calculating the reciprocal of the arithmetic mean of the reciprocals of the values, and thus it is suitable for cases where proportionality and balance are key factors.

\begin{equation}
p^m = \frac{1}{\sum_{i\in C^r}{\frac{w_i}{p_i}}} \cdot \Sigma_{i\in C^r}{(w_i \cdot p_i)}.
\label{eqn_params}
\end{equation}


In the following rounds of federation, the normalized aggregated parameters  $p^m$ are extended as payout to the subsequent cohorts of collaborators.

\begin{algorithm}
\caption{SimHAgg aggregation algorithm}
\label{simagg_algo}
\begin{algorithmic}[1]
\Procedure{Weight Aggregation}{$C^r$, $p_{C^r}$}

    \State $\epsilon$ $\leftarrow$ $1e-5$ \Comment{$C^r$ = set of collaborators (at round $r$)}
    
    \State  $\hat{p}$ = average($p_{C^r}$) using \textbf{Eq.~\ref{eqn_pavg}} \Comment{$p_{C^r}$ = parameters of the collaborators in $C^r$}
    
    \For{$c$ in $C^r$} 
        \State Compute similarity weights $u_c$ using \textbf{Eqs.~\ref{eqn_sim}} and \textbf{\ref{eqn_sim_weights}}
        \State Compute sample weights $v_c$ using \textbf{Eq.~\ref{eqn_sample_weights}}
    \EndFor 
    
    \For{$c$ in $C^r$} 
        \State Compute aggregation weights $w_c$ using \textbf{Eq.~\ref{eqn_fsim_weights}}
    \EndFor  

    \State Compute master model parameters $p^m$ using \textbf{Eq.~\ref{eqn_params}}
    
    \State \textbf{return} $p^m$
\EndProcedure
\end{algorithmic}
\end{algorithm}

\section{Deep Learning Experiments}
\label{sec:results}

\subsection{Training Setup}
The experimental framework has been rigorously formulated to optimize the process of federated lesion segmentation, encompassing various methodological components. These components include streamlined aggregation methodologies, judicious selection of client participants, round-specific training protocols, and efficient strategies for communication. The overarching objective is to achieve an optimal selection of collaborating participants by means of RL strategies while ensuring the proficient aggregation of model updates obtained from specific contributors.

In this work, a comprehensive dataset comprising the records of 1251 patients, sourced from multiple institutions, was employed for training purposes. An additional set consisting of 219 patients was reserved to serve as a validation cohort. The collaborative aspect of this study involved the active participation of 33 individuals who played a pivotal role in the partitioning of the dataset.

The experimental setup featured the utilization of a preconfigured 3D U-shape neural network, focusing on semantic segmentation tasks pertaining to the delineation of the whole tumor (WT), tumor core (TC), and enhancing tumor (ET) regions. These tasks were executed through the utilization of Intel's OpenFL platform. The evaluation criteria encompassed DICE similarity and Hausdorff (95\%) distance, providing a robust framework for the assessment of the efficacy of the aggregation rounds, as elaborated upon in~\cite{pati2021federated}.

This experimental configuration serves as a robust foundation for the systematic evaluation of performance enhancements resulting from the proposed RL-based collaborator selection and model aggregation strategies within the overarching framework of FL. 

The hyperparameters used are shown in Table~\ref{turns_Hyperparameters}. 

\begin{table}[H]

\centering
\caption{Hyperparameters used in aggregation algorithms.}
\label{turns_Hyperparameters}
\resizebox{0.5\columnwidth}{!}{%
\begin{tabular}{ll}
\hline
Hyperparameter       & RL-HSimAgg \\  \hline
Learning rate        & 5e-5    \\
Epochs per round     & 1.0                                   \\
Communication rounds & 25                                  \\ \hline
\end{tabular}%
}
\end{table}

\subsection{Results}

In this section, we provide a brief summary of the collaborator selection process utilizing reinforcement techniques with the adaptation of modified similarity weighted aggregation (RL-HSimAgg). The results emphasize that the proposed methods exhibit swift convergence and sustained stability as the learning process progresses, encompassing all the assessed criteria.

\subsubsection{Model training and performance using internal validation data}
\begin{figure*}[htbp]
    \centering
    \begin{subfigure}[b]{0.45\textwidth}
        \includegraphics[width=\textwidth]{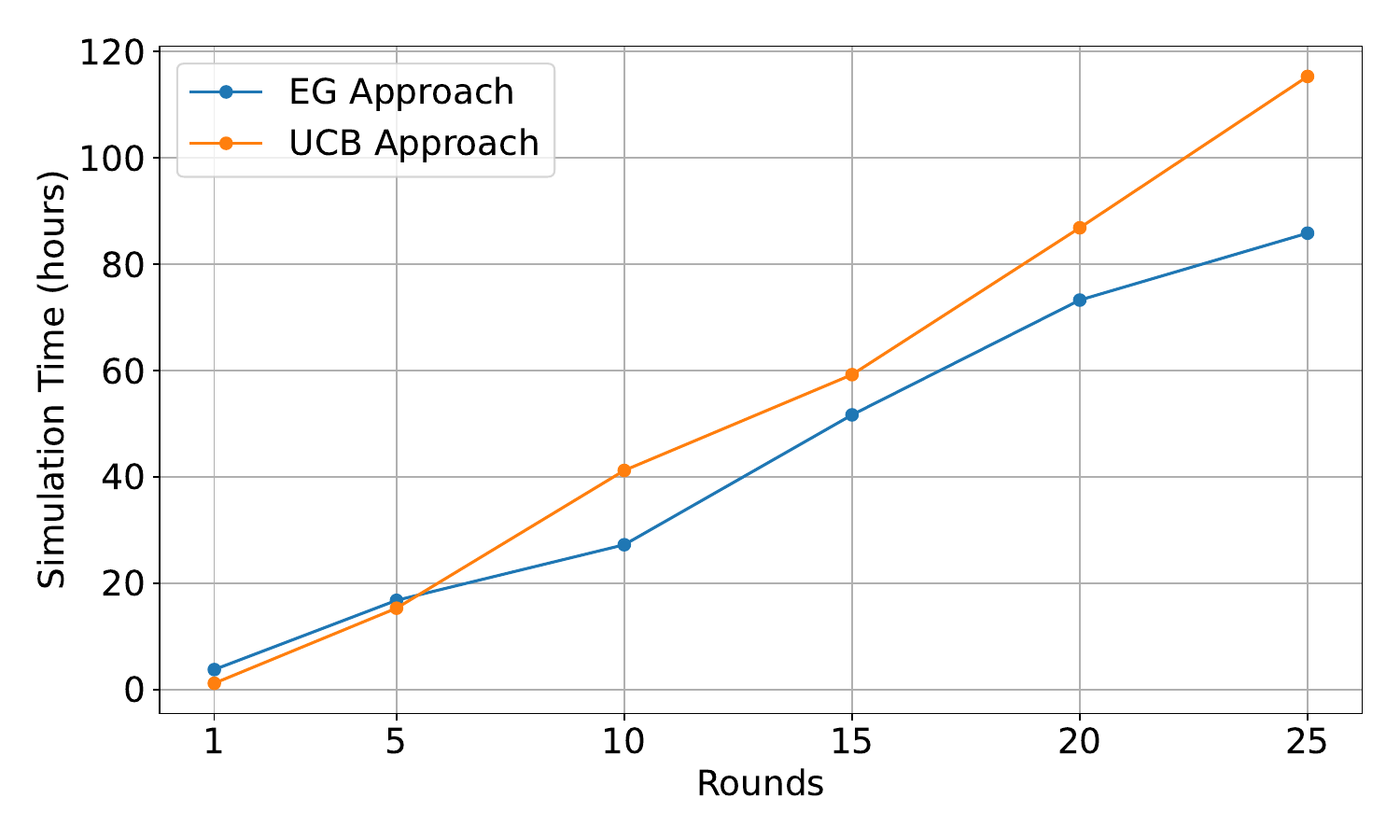}
        \caption{Simulation Time}
        \label{fig:sub1}
    \end{subfigure}
    \hfill
    \begin{subfigure}[b]{0.45\textwidth}
        \includegraphics[width=\textwidth]{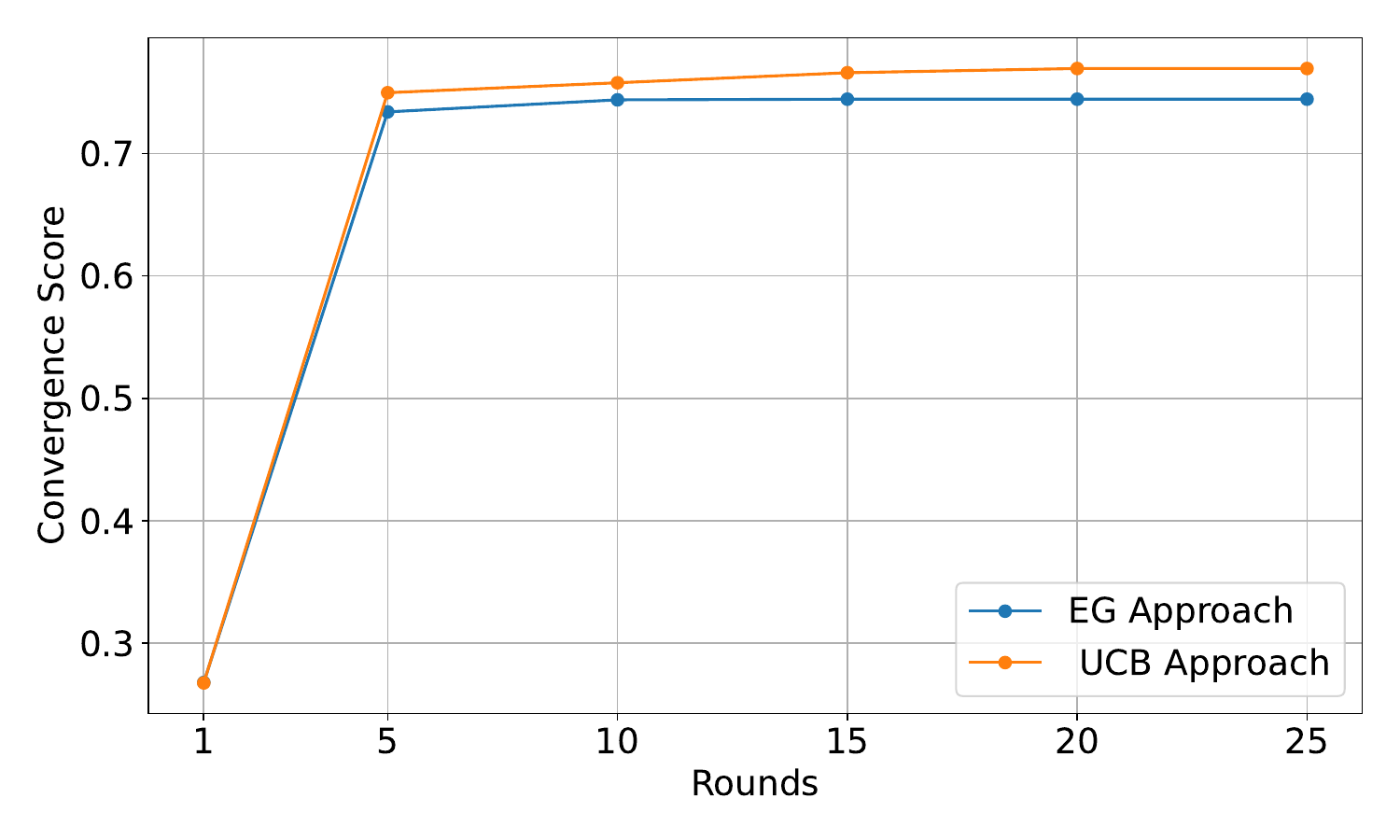}
        \caption{Projected Convergence Score}
        \label{fig:sub2}
    \end{subfigure}
    \\
    \begin{subfigure}[b]{0.45\textwidth}
        \includegraphics[width=\textwidth]{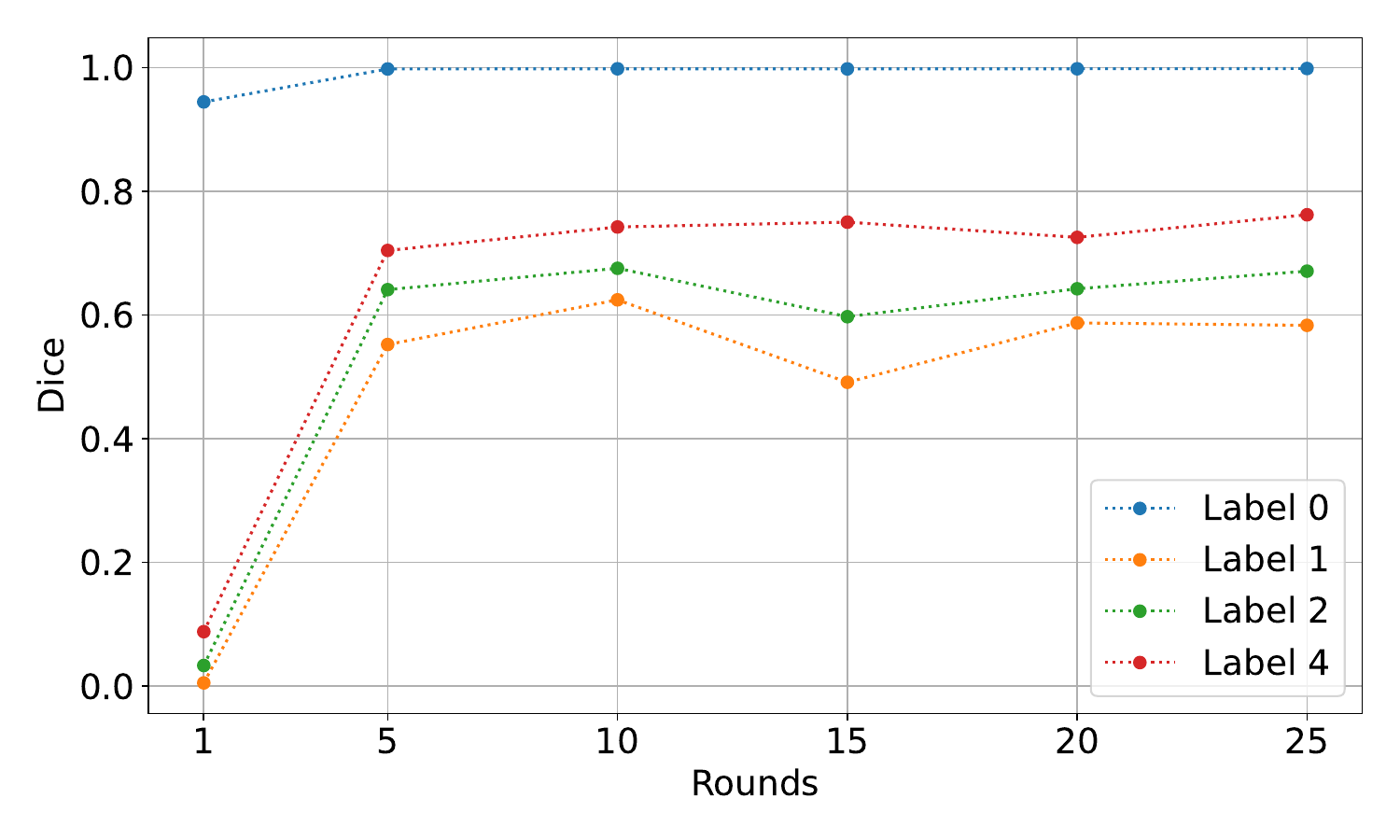}
        \caption{Dice similarity  (EG Approach)}
        \label{fig:sub3}
    \end{subfigure}
    \hfill
    \begin{subfigure}[b]{0.45\textwidth}
        \includegraphics[width=\textwidth]{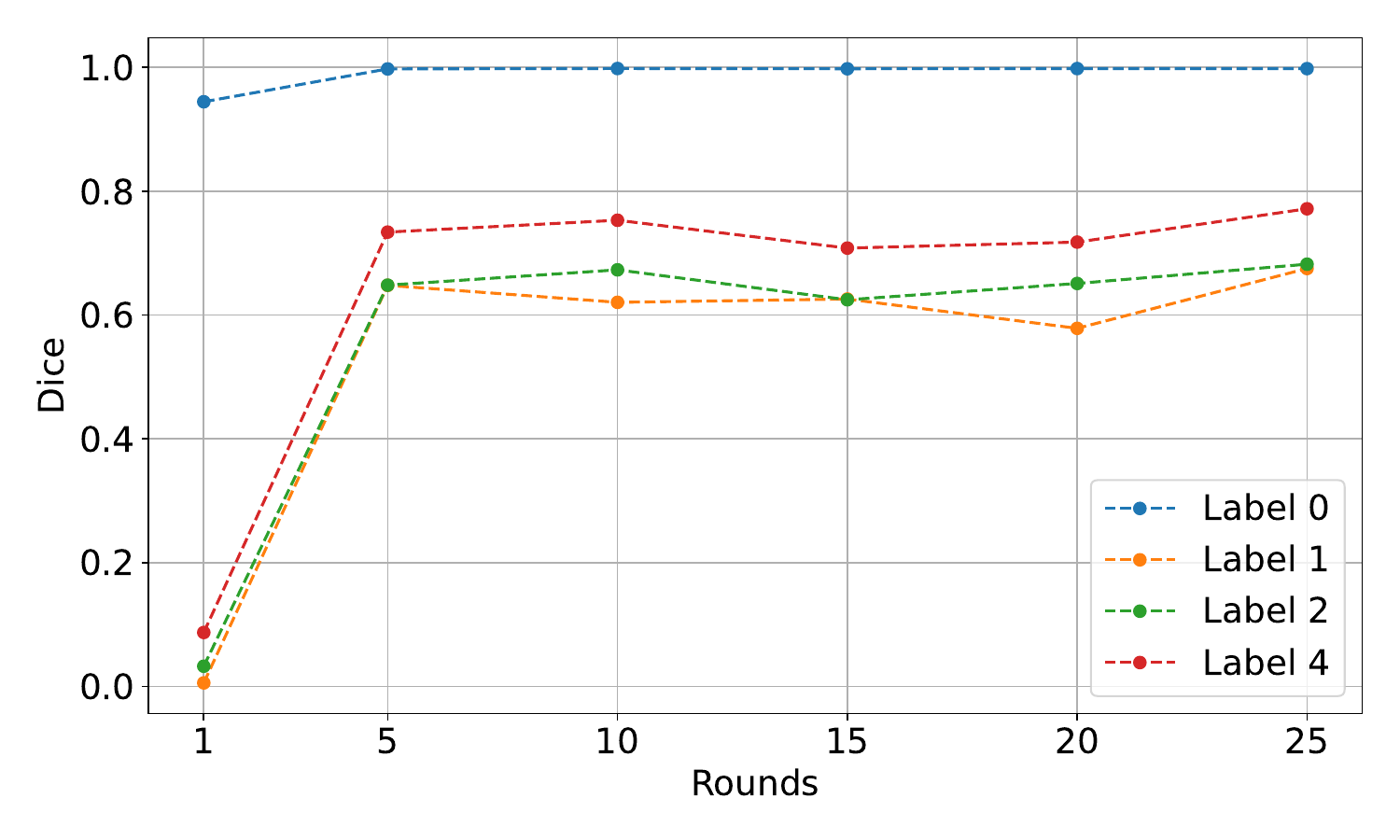}
        \caption{Dice similarity (UCB Approach)}
        \label{fig:sub4}
    \end{subfigure}
    \caption{Performance metrics for model training of RL-HSimAgg. The horizontal axis refers to the number of rounds and the vertical axis to the metrics.}
    \label{fig:foursubfigures}
\end{figure*}

Figure~\ref{fig:foursubfigures} shows the model training performance on internal validation data observed over 25 rounds of federated model training calculating simulated time, convergence score and DICE scores. UCB training took 20:38:50 wall-clock time and consumed 350.44 GB memory using 3.78 KWh energy on GPU, whereas, GE training took 15:34:49 wall-clock time and consumed 350.38 GB memory using 2.98 KWh energy on GPU. The labels utilized in the figure, namely 0, 1, 2, and 4, serve to delineate distinct classes relevant to brain tumor segmentation. Specifically, label 0 signifies normal brain tissue, label 1 represents the entire tumor along with the surrounding edematous area, label 2 corresponds to the tumor core, and label 4 is used to denote the enhancing region of the tumor.

\subsubsection{Model performance using external validation data}
We also assessed the performance of our approach using 219 instances of external validation data, as detailed in Table~\ref{tab:lesion} for the lesion-wise evaluations and Table~\ref{tab:brain} for the entire brain map. The lesion-wise Dice score and Hausdorff distance are metrics designed to assess a model's performance at the individual lesion level rather than the overall image level. This approach allows for a more nuanced understanding of how well models detect and delineate specific abnormalities. By evaluating each lesion separately, we can gain insight into a model's ability to identify and segment both small and large lesions accurately, avoiding bias towards models that excel only at capturing larger abnormalities. In general, UCB Approach exhibits superior performance in all segmentation tasks. The difference between the two approaches is notable in terms of Dice score. However, the UCB approach resulted in a smaller Hausdorff (95\%) as compared to the EG approach for all tumor regions. The UCB approach showed slower but higher convergence scores as well.

\begin{table}[th]
\centering
\caption{Lesion-wise performance evaluation on validation data.}
\label{tab:lesion}
\resizebox{0.8\columnwidth}{!}{%
\begin{tabular}{lcc}
\hline
Metrics & UCB Approach & EG Approach \\ \hline
Dice ET        & 0.4686139643 & 0.3275214475  \\
Dice TC        & 0.4675145428 & 0.3362944314  \\
Dice WT        & 0.2213302011 & 0.08511542852  \\
Hausdorff (95\%) ET & 161.4235281 & 219.8535584 \\
Hausdorff (95\%) TC & 153.7067903 & 212.7243962 \\
Hausdorff (95\%) WT & 274.9896873 & 337.833741 \\\hline
\end{tabular}%
}
\end{table}

\begin{table}[th]
\centering
\caption{Performance evaluation on the external validation data over the entire brain map.}
\label{tab:brain}
\resizebox{0.8\columnwidth}{!}{%
\begin{tabular}{lcc}
\hline
Metrics & UCB Approach & EG Approach \\ \hline
Dice ET        & 0.7334065378 & 0.6797028975  \\
Dice TC        & 0.7432201264 & 0.6821179569  \\
Dice WT        & 0.8252369483 & 0.7931141865  \\
Hausdorff (95\%) ET & 29.31827086 & 46.31837174 \\
Hausdorff (95\%) TC & 26.15306301 & 32.31553858 \\
Hausdorff (95\%) WT & 28.02696224 & 41.99560331 \\
Sensitivity ET & 0.7006077605 & 0.6841866812 \\
Sensitivity TC & 0.7075141203 & 0.662174806 \\
Sensitivity WT & 0.8106111042 & 0.8258257161 \\
Specificity ET & 0.9998288183 & 0.9995920256 \\
Specificity TC & 0.9998164178 & 0.9996753864 \\
Specificity WT & 0.9990476885 & 0.9984535062 \\ \hline
\end{tabular}%
}
\end{table}

\section{Discussion}
\label{sec:discussion}
This research work highlights the efficacy of incorporating RL strategies into the collaborator selection process. By harnessing the principles of EG and UCB algorithms, we have addressed the critical challenge of optimally selecting collaborators from a diverse pool of available collaborators for each round of FL.

Our findings demonstrate that the integration of RL algorithms enables the system to dynamically adapt and make intelligent decisions in real-time, leading to a more refined and efficient training process for the machine learning model. Currently, there is a limitation that data distributions and the number of available collaborators are fixed. Moreover, the research illustrates the significance of considering the inherent heterogeneity in collaborators contributions and data distributions, as reflected in non-IID data scenarios. By adopting multi-armed bandit-like strategies, we achieve a balance between exploration and exploitation, allowing for the identification and selection of optimal collaborators based on their performance histories. This approach cultivates an electoral environment wherein collaborators compete to participate in subsequent federation rounds, fostering healthy competition and enhancing the overall quality of the trained model.

In our current federated lesion segmentation experiments, two limitations are that the number of collaborators is fixed and the data distribution across collaborators remains static over rounds. To deal with this simplified scenario, we employed a straightforward alternating selection strategy that chooses the best and worst collaborators in even and odd rounds respectively. However, in a dynamic real-world setting, both the number of collaborators and the data distribution at each collaborator would vary over time. To adapt this selection strategy for such evolving conditions, we could employ more advanced multi-armed bandit algorithms that dynamically select collaborators to maximize the overall expected reward or utility in each round.

Specifically, bandit algorithms such as Thompson sampling or upper confidence bound algorithms could be applied to balance exploring new collaborators versus exploiting known high performers. The expected reward for collaborator selection can be defined using metrics such as model accuracy, data diversity, or other factors correlated with improved global model performance. Environment conditions like changing collaborator populations and data drift can be encoded in the prior selection for Bayesian approaches. This adaptive online optimization of collaborator selection based on changing utility and environment conditions can maximize overall learning performance in dynamic federated networks compared to fixed alternating selection rules. More research is still needed to develop such responsive selection algorithms for real-world FL.

The findings presented herein underscore the potential of RL algorithms in enhancing the FL process. As future work, we envision the exploration of advanced RL paradigms and the investigation of their application in diverse domains to further refine the collaborator selection process. We will explore alternative RL methodologies to effectively optimize collaborator selection for FL model training. These methodologies involve an algorithm that assesses the states of collaborators within the FL system and defines an action space to navigate to subsequent system states. In this context, a neural network, referred to as the Q-network, operates within the framework of Deep Q-Network (DQN) to estimate Q-values associated with a range of state-action pairs. These Q-values encapsulate the projected cumulative rewards achievable by taking an action from a given state. Throughout the training process, the DQN strives to minimize disparities between predicted Q-values and actual rewards earned. The transition between states is governed by the policy design, encompassing exploration strategies and recognized rewards. This state-to-state progression adheres to the guidance provided by the Bellman equation, a fundamental concept in RL that articulates the relationship between a state's value and the anticipated cumulative reward that can be obtained from that state onward. This methodology effectively balances immediate rewards with forecasts of forthcoming rewards \cite{Xu_2023}. With the proliferation of FL across various industries, our approach holds promise in bolstering the efficiency, robustness, and scalability of this decentralized learning paradigm.

\section{Conclusion}
\label{sec:conclusion}

This study demonstrates the effectiveness of integrating RL into the collaborator selection process in federated lesion segmentation, addressing the challenge of efficient collaborator selection. RL algorithms enable real-time, adaptive decision-making, enhancing model training. Despite limitations related to fixed data distributions and available collaborators, this approach remains valuable for scenarios involving evolving data and new collaborators, promoting equitable collaboration and improved model generalization. The study underscores the importance of accommodating diverse contributors and data distributions, particularly in non-IID contexts. Multi-armed bandit-like strategies strike a balance between exploration and exploitation, enhancing overall model quality through competitive collaborator selection. In conclusion, this research highlights RL's potential to advance FL in brain tumor segmentation and suggests future directions for refining collaborator selection methods in this evolving paradigm.

\section*{Acknowledgements}
This work was supported by the Business Finland under Grant 1337/31/2024. We also acknowledge the support and computational resources facilitated by the CSC-Puhti super-computer.

\bibliographystyle{splncs04}
\bibliography{references.bib}
\end{document}